\documentclass[runningheads]{llncs}

\usepackage{eccv}

\usepackage{eccvabbrv}

\usepackage{graphicx}
\usepackage{booktabs}
\usepackage{multirow}
\usepackage[accsupp]{axessibility}  

\usepackage{hyperref}

\usepackage{orcidlink}

\begin{document}

\title{Consistent Feature Transport for Image Relighting} 


\author{
Bohan Zhang\inst{1}$^\dagger$\orcidlink{0009-0006-1512-1405} \and
Huanwei Liang\inst{2}$^\dagger$\orcidlink{0009-0002-8490-2919} \and
Yuhan He\inst{2}\orcidlink{0009-0003-2725-9178} \and 
Hongteng Xu\inst{3}\orcidlink{0000-0003-4192-5360} \and 
Quxiao Chao\inst{2}\orcidlink{0009-0002-0247-2637} \and
Luoqi Liu\inst{2}\and 
Dixin Luo\inst{1}$^*$\orcidlink{0000-0003-1136-8903} \and
Ting Liu\inst{2}$^*$\orcidlink{0009-0008-7988-5935}
}

\authorrunning{B.~Zhang et al.}

\institute{School of Computer Science and Technology, Beijing Institute of Technology, Beijing 100081, China \and
MT Lab, Meitu Inc., Beijing 100083, China \and
Gaoling School of Artificial Intelligence, Renmin University of China, Beijing 100872, China}

\begingroup
\renewcommand\thefootnote{}
\footnotetext{$^\dagger$ Equal contribution. Work was done while Bohan Zhang was an intern at MT Lab, Meitu Inc., Beijing, China.}
\footnotetext{$^*$ Corresponding authors: dixin.luo@bit.edu.cn, lt@meitu.com.}
\endgroup

\maketitle

\begin{abstract}
Image relighting modifies illumination while preserving non-lighting content such as identity and geometry. 
Existing diffusion-based methods often suffer from unstable illumination changes or inconsistent content preservation under complex lighting, as they lack an explicit mechanism to learn feature transformations between images.
We reformulate relighting as an illumination feature transport problem and introduce Consistent Feature Transport (CFT), a training principle that explicitly enforces illumination-consistent transport between source and target image distributions. 
Built upon rectified flow, CFT jointly models noise-to-image generation and illumination-consistent source-to-target transport through trajectory-level supervision. 
This dual-transport formulation encourages isolation of illumination-specific variations while preserving content-aligned features.
To support complex lighting scenarios, we construct a large-scale portrait relighting dataset with diverse relighting effects. 
Experiments show consistent improvements over existing state-of-the-art relighting approaches and demonstrate that CFT can generalize to other editing tasks, including style transfer.
Code is available at \url{https://github.com/Dixin-Lab/CFT}.



\end{abstract}




\section{Introduction}

Image relighting aims to modify the illumination conditions of a given image while preserving its underlying content, including scene structure, identity, and material appearance. 
The task focuses on editing lighting attributes such as direction, intensity, color, and the interaction between light and other content elements. 
As a core problem in image editing, relighting has attracted increasing attention with the rapid development of generative models, particularly diffusion-based methods that enable flexible manipulation of visual attributes. 
Accurate and consistent relighting is essential for a range of applications, including portrait enhancement~\cite{kim2024switchlight,ren2024relightful}, virtual cinematography~\cite{zhang2025scaling,li2025translight}, and visual effects production~\cite{magar2025lightlab,chaturvedi2025synthlight}. 





With the advancement of diffusion-based image generation and editing methods, existing image relighting methods can be broadly categorized into three paradigms according to how they model illumination transformation. 
The first paradigm is control-based relighting, which formulates illumination editing as a conditional generation task.
These methods incorporate additional illumination-related signals~\cite{zhang2025scaling,kocsis2024lightit,liu2025dreamlight} to guide the diffusion process.
While such designs improve controllability, the final results are heavily dependent on the accuracy and completeness of the provided control signals. 
In practice, precisely specifying complex illumination effects remains challenging, which may limit robustness, consistency, and generalization.
The second paradigm is decomposition-based relighting, where illumination changes are achieved by manipulating intermediate representations~\cite{lyu2025intrinsicedit,zhang2024latent,xing2025luminet,li2025translight}. 
By operating in a decomposed space, these approaches attempt to isolate lighting factors from geometry and appearance. 
However, intrinsic decomposition itself may be inaccurate, especially under complex illumination conditions.
Errors in decomposition can propagate to the edited results, leading to inconsistent shading, color shifts, or loss of fine details.
The third paradigm formulates relighting as direct image-to-image translation, modeling the transport between source and target image distribution using generative models~\cite{chadebec2025lbm}.
This formulation treats relighting as an instance-level translation problem and does not explicitly model illumination-specific feature displacement. 
As a result, while structural fidelity can be maintained, the controllability and precision of illumination editing are limited.

Apart from methodological challenges, data availability also constrains the progress of image relighting.
While several open-source datasets exist for relighting objects or generic scenes~\cite{murmann2019dataset,hold2019deep,liu2023openillumination,helou2020vidit}, publicly available portrait relighting datasets remain limited. 
Existing portrait relighting data~\cite{debevec2000acquiring,pandey2021total,kim2024switchlight,wang2025comprehensive} often lack sufficient diversity in lighting configurations, particularly scenarios involving complex illumination effects, which hinders the development and evaluation of methods designed for realistic and controllable image relighting.




In this paper, we reformulate relighting as an illumination feature transport problem and introduce \textbf{C}onsistent \textbf{F}eature \textbf{T}ransport (CFT) for image relighting, which explicitly models the illumination-consistent transport between source and target image distributions. 
Specifically, we build our method upon rectified flow and consider the relationships among the noise, source image, and target image distributions. 
Similar to conventional image generation and editing frameworks, we model both noise-to-source and noise-to-target transport. 
For the source-to-target transport, instead of directly supervising with the original image pairs, we use alternative pairs that share the same illumination condition as supervision. 
This design encourages the model to focus on an illumination feature transformation rather than a generic image-to-image modification.
By jointly modeling noise-to-image generation and illumination-consistent source-to-target transport, the framework guides the generative process to focus on transporting illumination features while preserving identity, geometry, and other non-lighting attributes.
In addition, we construct a dedicated portrait relighting dataset to support learning under complex lighting conditions. 
The dataset contains diverse portrait images under a wide range of illumination effects, providing richer supervision for modeling lighting variations.


In summary, this work makes the following contributions. 
First, we introduce a consistent feature transport principle for image relighting that explicitly models the illumination-consistent transport between the source and target image distributions.
The proposed framework encourages faithful preservation of image content while improving targeted modification of lighting features. 
Second, we present a portrait relighting dataset with diverse light configurations and complex illumination effects.
This dataset fills the gap in existing resources, where portrait data with complex illumination is scarce.
Third, extensive experiments demonstrate the effectiveness of the proposed approach, yielding consistent improvements over existing baselines. 
Beyond relighting, we further show that the proposed method can generalize to other editing tasks, such as style transfer, highlighting its potential as a unified strategy for image editing.

\section{Related Work}

\subsection{Diffusion-based Image Editing}

Diffusion-based image editing aims to modify specific visual attributes while preserving the underlying content of the input image. 
Existing approaches differ primarily in how the source image and desired modification are incorporated into the generative process.
A common paradigm is control-based generation, where the model starts from noise and generates the edited target image conditioned on the source image and additional control signals, such as textual instructions~\cite{brooks2023instructpix2pix,zhang2023magicbrush}, reference images~\cite{wang2025omnistyle,yang2023imagebrush,choi2024improving}, or disentangled representations~\cite{dalva2025fluxspace,zhang2024choose,pan2025counterfactual}. 
In this formulation, editing is realized as conditional generation. 
While effective, the quality of editing largely depends on the expressiveness and accuracy of the control conditions.
Another line of work adopts trajectory-based editing, which manipulates diffusion paths to steer the model from the source image toward the edited target result.
Both training-based approaches~\cite{chadebec2025lbm,tan2025vision,chen2026car} and training-free methods~\cite{wang2025taming,deng2025fireflow,cao2023masactrl,kulikov2025flowedit,kim2026flowalign} operate directly on the diffusion trajectory, starting from the source image and progressively guiding it toward the desired target. 
While such designs help maintain structural consistency due to the strong anchoring effect of the source image, they typically focus on instance-level alignment or generic domain transport. 
As a result, the editing flexibility is often limited, and achieving substantial or fine-grained attribute modifications can be challenging.

Overall, existing diffusion editing frameworks either rely on conditional generation or trajectory manipulation, but rarely consider explicit and direct feature transport between source and target domains.

\subsection{Image Relighting Methods}

Image relighting aims to selectively modify illumination without altering other content, making it particularly sensitive to how transformations are modeled.
Early neural relighting methods were largely grounded in physically inspired modeling. 
These approaches leveraged explicit illumination representations~\cite{li2022physically,zhang2024latent}, intrinsic decomposition~\cite{yu2020self,kocsis2024intrinsic,lyu2025intrinsicedit}, or physics-guided reflectance models~\cite{kim2024switchlight,pandey2021total} for lighting variations. 
By introducing physical priors, they enabled interpretable control over light sources and shading behavior. 
However, such methods often depend on accurate intrinsic decomposition or specialized setups, which limit their ability to handle images with complex illumination effects.

With the emergence of diffusion models, relighting has increasingly been addressed within a generative editing framework.
Following the control-based paradigm, several works introduce illumination-related conditions, such as environmental maps~\cite{zhang2025scaling,chaturvedi2025synthlight}, intrinsic components~\cite{kocsis2024lightit,magar2025lightlab,choi2025scribblelight,liu2026unilumos}, or lighting representations~\cite{liu2025dreamlight,bharadwaj2025genlit}, to guide the denoising process. 
Although this improves controllability, the effectiveness of relighting depends heavily on the quality and completeness of the control signals. 
Precisely specifying complex illumination effects, especially under multiple light sources or spatially varying shadows, remains challenging.
Alternatively, relighting has also been formulated as image-to-image or trajectory-based translation, where the model directly steers the diffusion path from the source image toward a target illumination domain. 
Recent works explore bridge-based formulations to learn mappings between image distributions~\cite{chadebec2025lbm}.
This paradigm treats relighting as an instance-level translation problem, which helps preserve the underlying content of the source image.
However, the direct instance-level translation also limits their ability to perform fine-grained and expressive illumination editing.

Overall, existing relighting approaches either depend on explicit illumination representations and conditions or perform instance-level diffusion transport. 
In contrast, our approach explicitly models how illumination features should be transported between source and target representations, enabling more reliable and consistent relighting under diverse and complex lighting conditions.

\section{Methodology}

\subsection{Proposed Method}

Let $x_{\mathrm{src}}$ denote the source image before relighting and $x_{\mathrm{tgt}}$ denote the corresponding relighting target image. 
The desired illumination change is specified by a conditioning signal $c$, represented as a textual description of the lighting transformation. 
The goal is to learn a conditional velocity model $v_{\theta}$ that generates the relighting image starting from noise while preserving the non-lighting content of $x_{\mathrm{src}}$. 
Following the rectified flow formulation~\cite{lipman2023flow,liu2023flow}, the model is trained to predict the transport velocity from a prior sample toward the target latent representation under the guidance of the source image and the textual condition. 
The training objective is defined as:
\begin{equation}
\label{equ:problem_formulation}
    \mathcal{L}_{\mathrm{relight}}(\theta) =
    \mathbb{E}_{z_0 \sim \pi_0,\; z_{\mathrm{tgt}},\; t \sim \mathcal{U}(0,1)}
    \left[
    \left\| v_{\theta}(z_t, t, x_{\mathrm{src}}, c) - (z_{tgt} - z_0) \right\|^2
    \right],
\end{equation}
where $z_0$ denotes a sample drawn from the prior distribution and $z_{tgt}$ denotes the latent representation of $x_{\mathrm{tgt}}$.

\begin{figure}[t]
    \centering
    \includegraphics[width=0.55\linewidth]{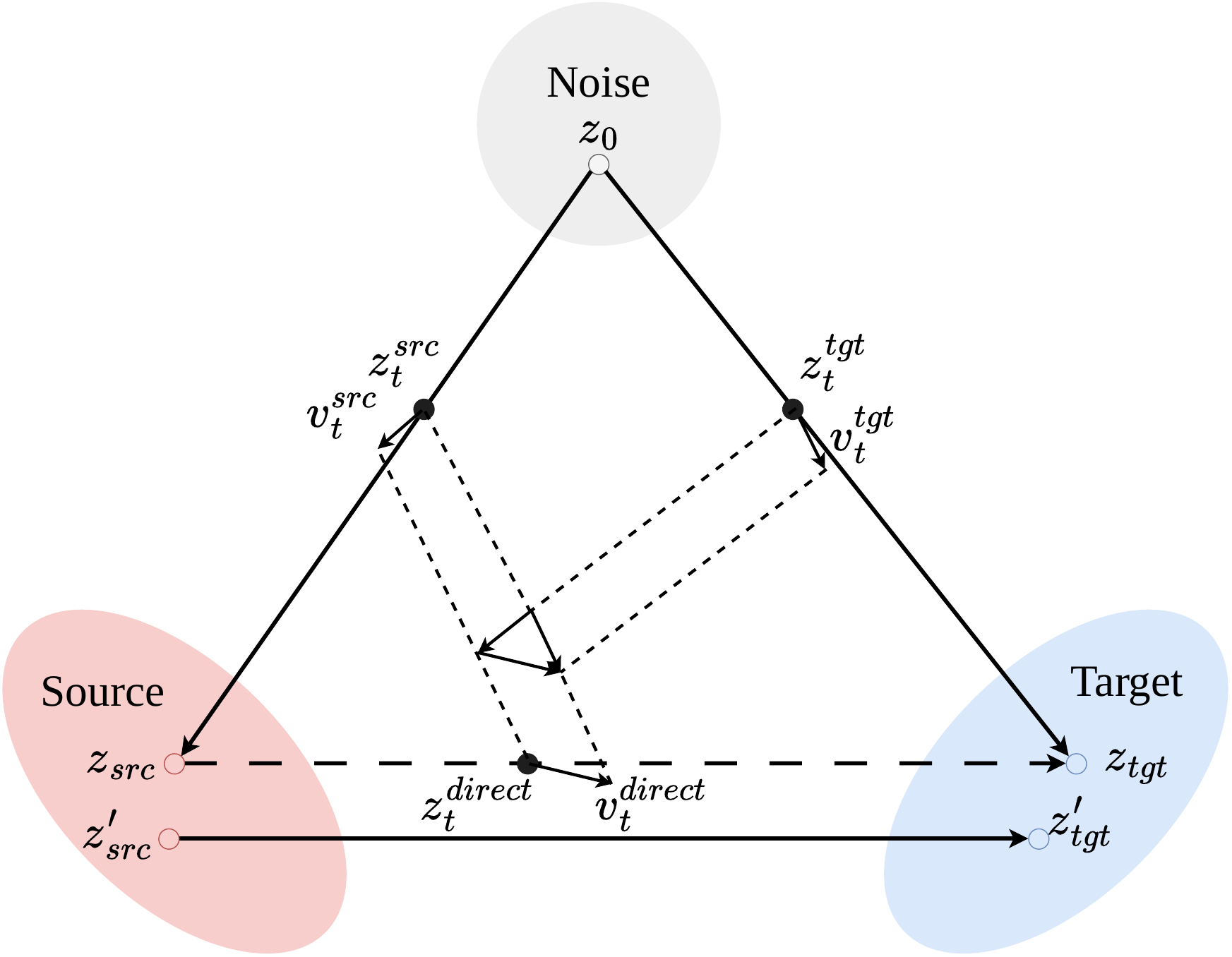}
    \caption{Overall framework of the proposed method.}
    \vspace{-1em}
    \label{fig:Overview}
\end{figure}

As illustrated in \Cref{fig:Overview}, we consider the relationships between the noise, source image, and target image distributions. 
Following the original rectified flow formulation, we train the model to learn the transport from Gaussian noise to the target image distribution. 
Specifically, given a source image $x_{\mathrm{src}}$, a target relighting image $x_{\mathrm{tgt}}$, and the corresponding relighting condition $c_{\mathrm{tgt}}$, the model is trained to predict the velocity field that transports samples from the prior toward the target latent representation:
\begin{equation}
    \mathcal{L}_{1} =
    \mathbb{E}_{z_0 \sim \pi_0,\; z_{\mathrm{tgt}},\; t \sim \mathcal{U}(0,1)}
    \left[
    \left\| v_{\theta}(z_t^{\mathrm{tgt}}, t, x_{\mathrm{src}}, c_{\mathrm{tgt}})
    - (z_{\mathrm{tgt}} - z_0) \right\|^2
    \right].
\end{equation}

To strengthen content preservation, we additionally train the model to learn the transport from noise to the source image distribution.
This objective corresponds to a reconstruction pathway under a neutral condition $c_{\mathrm{src}}$ from the same prior $z_0$:
\begin{equation}
    \mathcal{L}_{2} =
    \mathbb{E}_{z_0 \sim \pi_0,\; z_{\mathrm{src}},\; t \sim \mathcal{U}(0,1)}
    \left[
    \left\| v_{\theta}(z_t^{\mathrm{src}}, t, x_{\mathrm{src}}, c_{\mathrm{src}})
    - (z_{\mathrm{src}} - z_0) \right\|^2
    \right].
\end{equation}

The objectives $\mathcal{L}{1}$ and $\mathcal{L}{2}$ are sufficient for conditional generation, as they steer the model to generate the source and target image under different conditioning signals.
However, such conditional generation is insufficient for the control over illumination variation and often fails to produce consistent lighting transformations across different instances.
Therefore, beyond the noise-to-image generation objective, we explicitly model the transport between the source and target distributions, encouraging direct illumination transformation.

Following the inversion-free image editing methods~\cite{kulikov2025flowedit,deng2025fireflow}, we leverage the linearity of the rectified flow and approximate the intermediate states along the direct path via the parallelogram law:
\begin{equation}
    z_t^{\mathrm{direct}} = z_{\mathrm{src}} + z_t^{\mathrm{tgt}} - z_t^{\mathrm{src}}.
\end{equation}

According to the ODE formulation, the corresponding velocity along this path can be expressed as:
\begin{equation}
\label{equ:v_direct}
    v_t^{\mathrm{direct}} =
    v_{\theta}(z_t^{\mathrm{tgt}}, t, x_{\mathrm{src}}, c_{\mathrm{tgt}})
    - v_{\theta}(z_t^{\mathrm{src}}, t, x_{\mathrm{src}}, c_{\mathrm{src}}).
\end{equation}

During training, a straightforward approach is supervising $v_t^{\mathrm{direct}}$ with the original pair $(x_{\mathrm{src}},x_{\mathrm{tgt}})$.
However, since $v_t^{direct}$ is derived from the parallelogram law, it essentially models the transport from $x_{\mathrm{src}}$ to $x_{\mathrm{tgt}}$.
Directly using the pair $(x_{\mathrm{src}},x_{\mathrm{tgt}})$ as ground truth does not encourage the model to learn illumination features.
Instead, such supervision may encourage the model to fit sample-specific differences that entangle illumination variations with other non-lighting content.
Consequently, the model may rely on spurious content-dependent correlations to accomplish illumination edits, rather than learning illumination-specific transformations.

In contrast to the choice of the original pair $(x_{\mathrm{src}},x_{\mathrm{tgt}})$ for supervision, we apply another image pair $(x_{\mathrm{src}}', x_{\mathrm{tgt}}')$ that shares the same illumination transformation but differs in identity and scene content from the original image pair $(x_{\mathrm{src}}, x_{\mathrm{tgt}})$ as the ground-truth pathway. 
We can obtain the following objective:
\begin{equation}
\mathcal{L}_{3}
=
\mathbb{E}_{z_{\mathrm{src}}',\, z_{\mathrm{tgt}}',\, t}
\left[
\left\|
v_t^{\mathrm{direct}}
-
(z_{\mathrm{tgt}}' - z_{\mathrm{src}}')
\right\|^2
\right],
\end{equation}
where $v_t^{\mathrm{direct}}$ is defined in \Cref{equ:v_direct}. 
Thus, the model can learn illumination-consistent transport between the source and target image distributions, rather than fitting content-specific differences between individual image pairs.
The model is encouraged to represent relighting as a transport of illumination features, instead of a generic image-to-image modification.

The final training objective combines the above components:
\begin{equation}
    \mathcal{L} = \mathcal{L}_1 + \mathcal{L}_2 + \alpha \cdot \mathcal{L}_3,
\end{equation}
where $\alpha$ controls the strength of the consistent feature transport. 
This formulation enables the model to learn both noise-to-image generation and illumination-consistent feature transport, improving consistency of relighting and the overall generated result.

\begin{figure}[t]
    \centering
    \includegraphics[width=\linewidth]{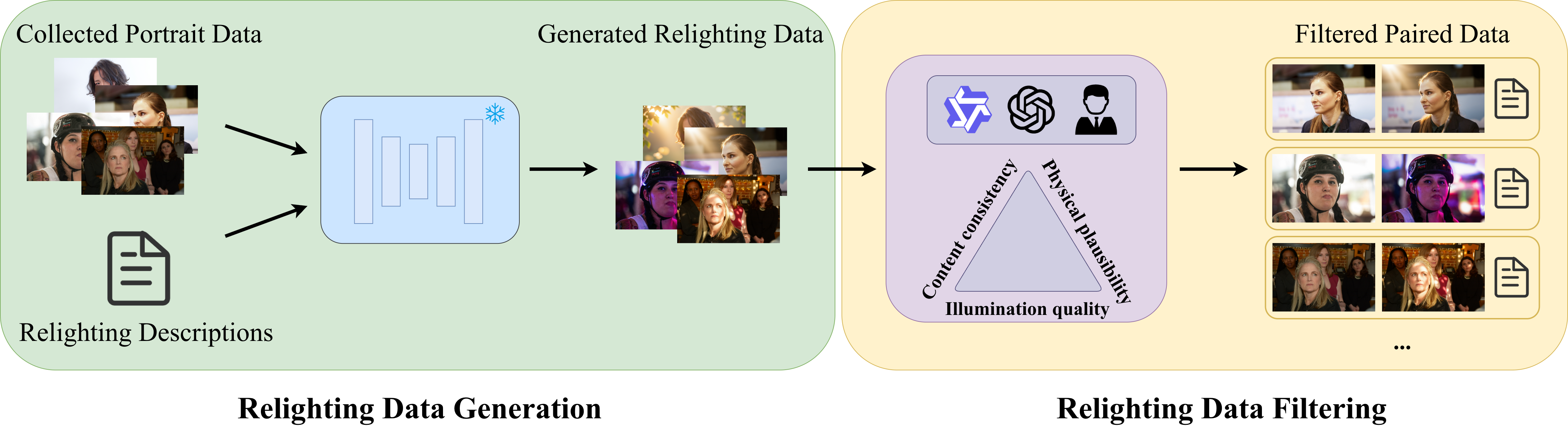}
    \caption{Dataset Construction Pipeline.}
    \vspace{-1em}
    \label{fig:data_construction_pipeline}
\end{figure}

\begin{figure}[t]
    \centering
    \begin{subfigure}{0.31\textwidth} 
     \includegraphics[width=\linewidth]{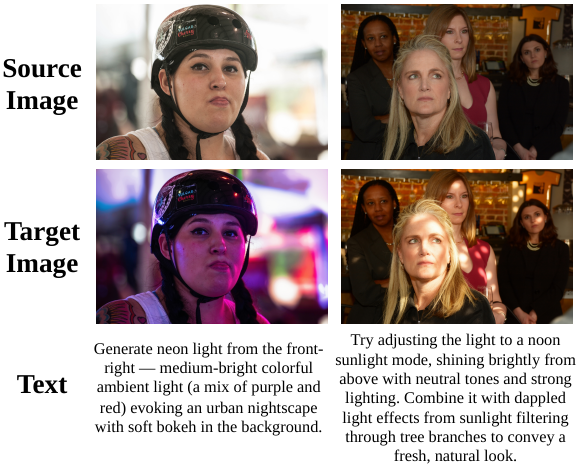}
    \caption{Examples of the constructed dataset}
    \label{fig:dataset_visualization}
  \end{subfigure}
  \hfill
   \begin{subfigure}{0.26\textwidth} 
    \includegraphics[width=\linewidth]{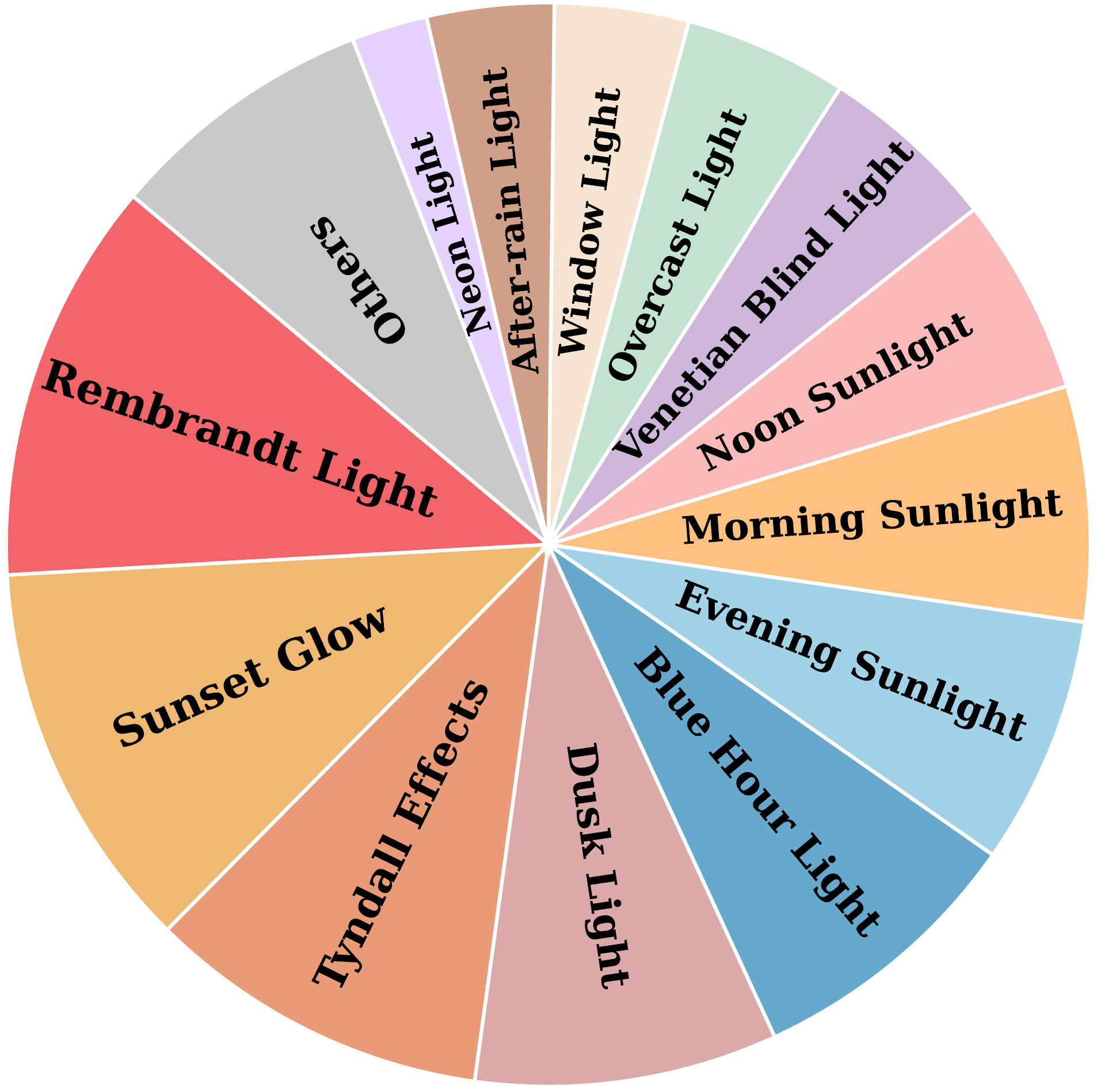}
    \caption{Distribution of light effect categories}
    \label{fig:light_effect_distribution}
  \end{subfigure}
  \hfill
   \begin{subfigure}{0.4\textwidth} 
    \includegraphics[width=\linewidth]{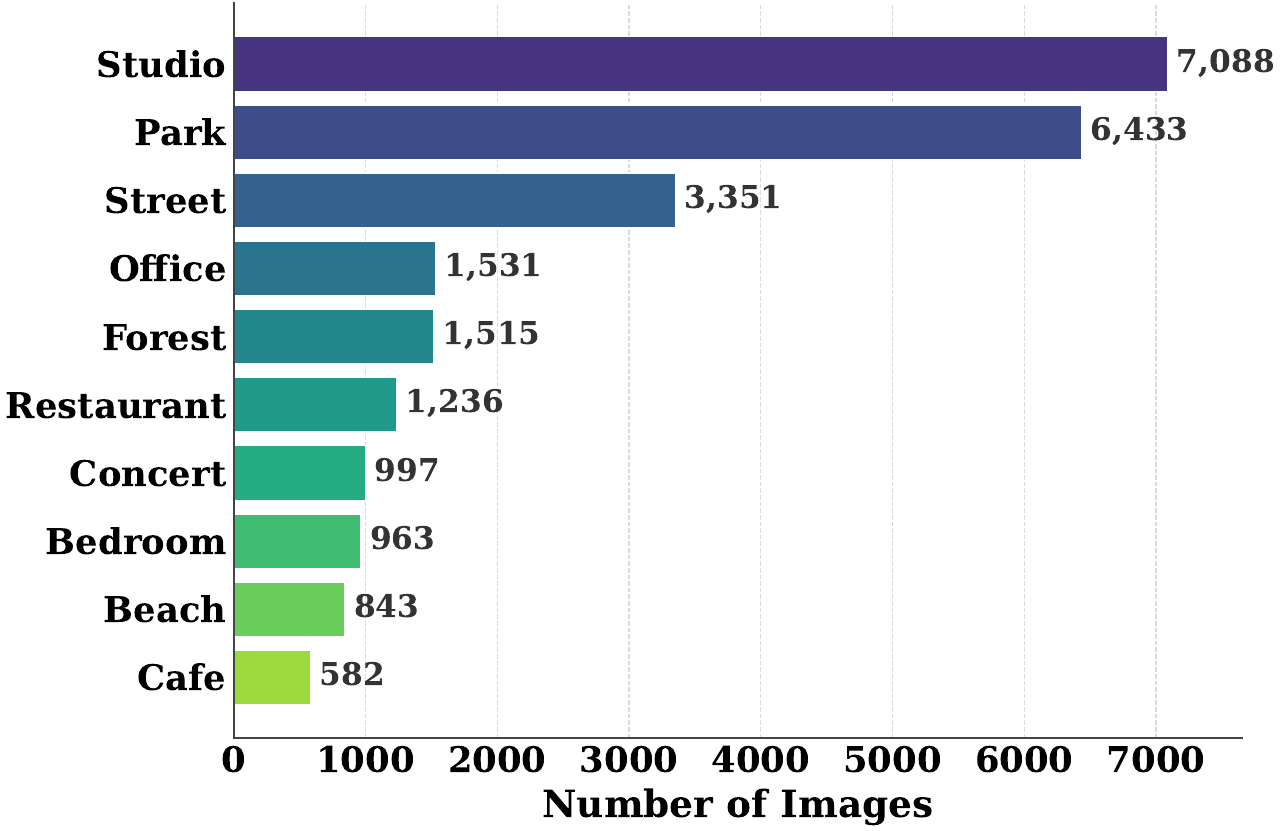}
    \caption{Top-10 scene categories}
    \label{fig:environment_type_top10}
  \end{subfigure}
  \caption{Overview of the constructed dataset.}
  \vspace{-1em}
  \label{fig:dataset_overview}
\end{figure}

\subsection{Dataset Construction}
Existing relighting datasets~\cite{murmann2019dataset,hold2019deep,liu2023openillumination,helou2020vidit} primarily focus on objects or general scenes, while publicly available portrait relighting datasets remain scarce. 
Moreover, many existing portrait relighting data~\cite{debevec2000acquiring,pandey2021total,kim2024switchlight,wang2025comprehensive} are captured under controlled environments with relatively simple illumination setups. 
Therefore, they lack diverse complex lighting effects and multiple light configurations, such as different colors, spatially varying illumination, and non-uniform shadows. 
This limitation reduces their effectiveness for training and evaluating models designed for realistic and controllable image relighting.

To address this gap, we construct a portrait relighting dataset with diverse and complex lighting effects.
The overall data construction pipeline is illustrated in \Cref{fig:data_construction_pipeline}.
First, we collect source images without explicit light effects from existing open-source portrait datasets~\cite{karras2019style,kalantari2017deep,hasinoff2016burst}.
These images exhibit substantial diversity in subjects, appearances, poses, and background environments, providing a rich set of base images for relighting data generation.
Meanwhile, we generate a comprehensive set of relighting descriptions that cover a broad spectrum of illumination styles.
We consider two major categories of lighting sources: natural lighting (\eg, sunlight, moonlight) and artificial lighting (\eg, streetlights, neon lights). 
For each category, the descriptions are constructed by specifying several key aspects of the lighting: 
(1) Light direction: describing the spatial direction of illumination, including front, side, and bottom lighting, as well as their combinations, such as front-left and back-right;
(2) Light configuration: representing typical lighting setups and effects, such as Tyndall light, Rembrandt lighting, and Blue Hour illumination;
(3) Photometric properties: characterizing lighting attributes including color temperature, light intensity, shadow softness, and overall atmospheric tone;
These aspects ensure the description spans both global illumination changes and spatially structured lighting patterns.
We then employ existing pre-trained models (Nano Banana, Seedream, and RoboNeo) to generate candidate relighting images under those descriptions.

Since automatically generated relighting results may vary in quality, we further employ an ensemble-based filtering pipeline to ensure dataset reliability. 
Specifically, we evaluate each source and target image pair using three models: QwenVL~\cite{bai2025qwen3}, EditScore~\cite{luo2026editscore}, and GPT-4o~\cite{hurst2024gpt}.
The evaluation focuses on three criteria: 
(1) Illumination quality: evaluating whether the generated lighting effects are visually plausible and consistent with the specified relighting descriptions;
(2) Content consistency: ensuring that non-illumination content, including facial identity, pose, and background, remains unchanged between the source and relit images; 
(3) Physical plausibility: assessing whether the lighting configuration satisfies basic physical constraints, such as consistent shading direction and coherent shadow patterns.
When all three models confirm the satisfaction of the three criteria, the image pair is retained in the dataset.
After filtering with three models, we further perform manual selection based on the above criteria to ensure data quality.

Finally, we construct a portrait relighting dataset consisting of 34,695 image pairs in total, including 34,249 pairs for training and 446 pairs for testing. 
\Cref{fig:dataset_overview} presents visual examples and statistics of the constructed dataset.
For illumination effects, we summarize the light-effect categories contained in the relighting descriptions and visualize their distribution by image count, as shown in \Cref{fig:light_effect_distribution}.
For the scenes in the dataset, indoor scenes account for 48.1\% and outdoor scenes account for 51.9\%.
We further perform a statistical analysis of the scene categories in the dataset. 
\Cref{fig:environment_type_top10} presents the number of images for the Top-10 scene categories.
The resulting dataset provides diverse light configurations and complex illumination effects.

\section{Experiments}

\subsection{Setup}
We compare our method with several representative approaches. 
(1) Control-based editing (Flux-Kontext~\cite{labs2025flux}, Qwen-Image-Edit-2509~\cite{wu2025qwen}, and IC-light~\cite{zhang2025scaling}): leveraging the diffusion-based image editing models with controllable signals as guidance for the image relighting task.
Notably, IC-light applies an environment map as a training condition, whereas it can perform inference without any additional control signals.
(2) Decomposition-based editing (Intrinsic Edit~\cite{lyu2025intrinsicedit} and Latent Intrinsic~\cite{zhang2024latent}): decomposing images into intrinsic components (\eg, shadings, albedos) or latent representations and performing relighting by editing those intrinsics.
(3) Image-to-image translation (LBM~\cite{chadebec2025lbm}, FlowEdit~\cite{kulikov2025flowedit}): performing editing directly through the diffusion path from the source image to the target image.
(4) Original pretrained models (Flux-Kontext, Nano Banana): directly using the pretrained model without fine-tuning on the relighting dataset.
Since IC-Light does not release training code, we reimplement the method for our experiments.

We evaluate relighting performance using four categories of metrics: pixel-level fidelity, perceptual similarity, distribution-level realism, and lighting-specific evaluation.
For pixel-level fidelity, Peak Signal-to-Noise Ratio (PSNR) and Structural Similarity Index (SSIM) are used to measure reconstruction accuracy and structural consistency of relighting images.
For perceptual similarity, Learned Perceptual Image Patch Similarity (LPIPS) is adopted to quantify perceptual differences in feature space, reflecting human-perceived visual similarity.
For distribution-level realism, Fréchet Inception Distance (FID) is employed to evaluate the realism of generated images by measuring the distributional distance between generated samples and ground-truth target images.
For lighting-specific evaluation, Estimated Irradiance (EI) is leveraged to compute the MAE between the predicted and ground-truth irradiance maps.

We implement Flux-Kontext and Qwen-Image-Edit-2509 as two base models for our proposed method.  
We set the hyperparameter $\alpha=0.1$.
We use a learning rate of $1\times10^{-4}$ and a batch size of 4. 
We train the models for 4{,}000 steps on 8 NVIDIA H20 GPUs.
Except for the training-free baselines and the original pretrained models, we train all the methods on our constructed dataset.
Unless otherwise specified, all experiments follow the same training configuration to ensure fair comparison across methods.


\begin{table}[t]
\caption{Quantitative comparison results.}
\label{tab:main-result}
\centering
\begin{tabular}{cccccc}
\hline
Methods                   & SSIM $\uparrow$           & PSNR $\uparrow$            & LPIPS $\downarrow$          & FID  $\downarrow$      & IE $\downarrow$      \\ \hline
IC-light                   & 0.7436 & 15.3314 & 0.2209 & 47.3196 & 15.3575 \\
LBM                       & 0.7908 & 15.5954 & 0.2267 & 40.8663 & 16.9276 \\
FlowEdit                  & 0.7133 & 13.6775 & 0.3298 & 78.1236 & 23.2227 \\
Intrinsic Edit            & 0.7148 & 15.4219 & 0.2646 & 57.3994 & 14.2292 \\
Latent Intrinsic          & 0.8425 & 18.1809 & 0.2379 & 35.0617 & 16.2387 \\
Nano Banana               & 0.7339 & 14.9196 & 0.2507 & 46.5012 & 15.9686 \\
Flux-Kontext (original)   & 0.6386 & 14.1543 & 0.2700 & 53.3793 & 18.2834 \\
Qwen-Image-Edit (w.o. CFT) & 0.882  & 20.117  & 0.114  & 30.2147 & 11.5505 \\
Flux-Kontext (w.o. CFT)   & 0.9153 & 22.9192 & 0.0985 & 20.3817 & 7.8155  \\ \hline
Qwen-Image-Edit (w. CFT)  & 0.8894 & 20.9176 & 0.1132 & 25.5529 & 10.8670 \\
Flux-Kontext (w. CFT)     & 0.9202 & 23.5105 & 0.0946 & 18.7338 & 7.4231  \\ \hline
\end{tabular}
\end{table}

\subsection{Main Results}

\Cref{tab:main-result} presents quantitative comparison results with representative relighting approaches. 
Compared with control-based editing baselines, CFT consistently improves performance when applied to both Flux-Kontext and Qwen-Image-Edit.  
By explicitly learning the source-to-target transport pathway during training, the model better captures illumination-consistent features while preserving other non-lighting content, such as identity and scene structure.
The improvement on both Flux-Kontext and Qwen-Image-Edit also demonstrates the general applicability of CFT to flow-based image editing models.
Moreover, our method also substantially outperforms IC-light.
IC-light requires additional control signals, such as environment maps, during training.
Constructing such auxiliary conditions is often labor-intensive and time-consuming. 
Our method directly learns illumination feature transport from paired data, leading to a more efficient training process.
Compared with decomposition-based editing methods (Intrinsic Edit and Latent Intrinsic), control-based editing approaches generally achieve stronger performance.
This may be attributed to the inaccuracy in the intrinsic decomposition and lighting-related intrinsic editing, which can propagate errors to the final relighting results, leading to inconsistent shading, color artifacts, and loss of fine illumination details. 
Compared with image-to-image translation methods (LBM, FlowEdit), control-based editing approaches also demonstrate superior performance. 
Instance-level translation does not explicitly model illumination-specific feature transformations, which may lead to suboptimal results, especially for images with complex content or structures. 
Compared with the original pretrained model (Flux-Kontext, Nano Banana), control-based editing approaches consistently exhibit better performance. 
This observation highlights the importance of training on relighting-specific data, as pretrained parameters alone are not sufficient to effectively handle relighting tasks.
Overall, the results validate the effectiveness of our proposed method and demonstrate the strong generalization of CFT across various image editing models.

\begin{figure}[t]
    \centering
    \includegraphics[width=\linewidth]{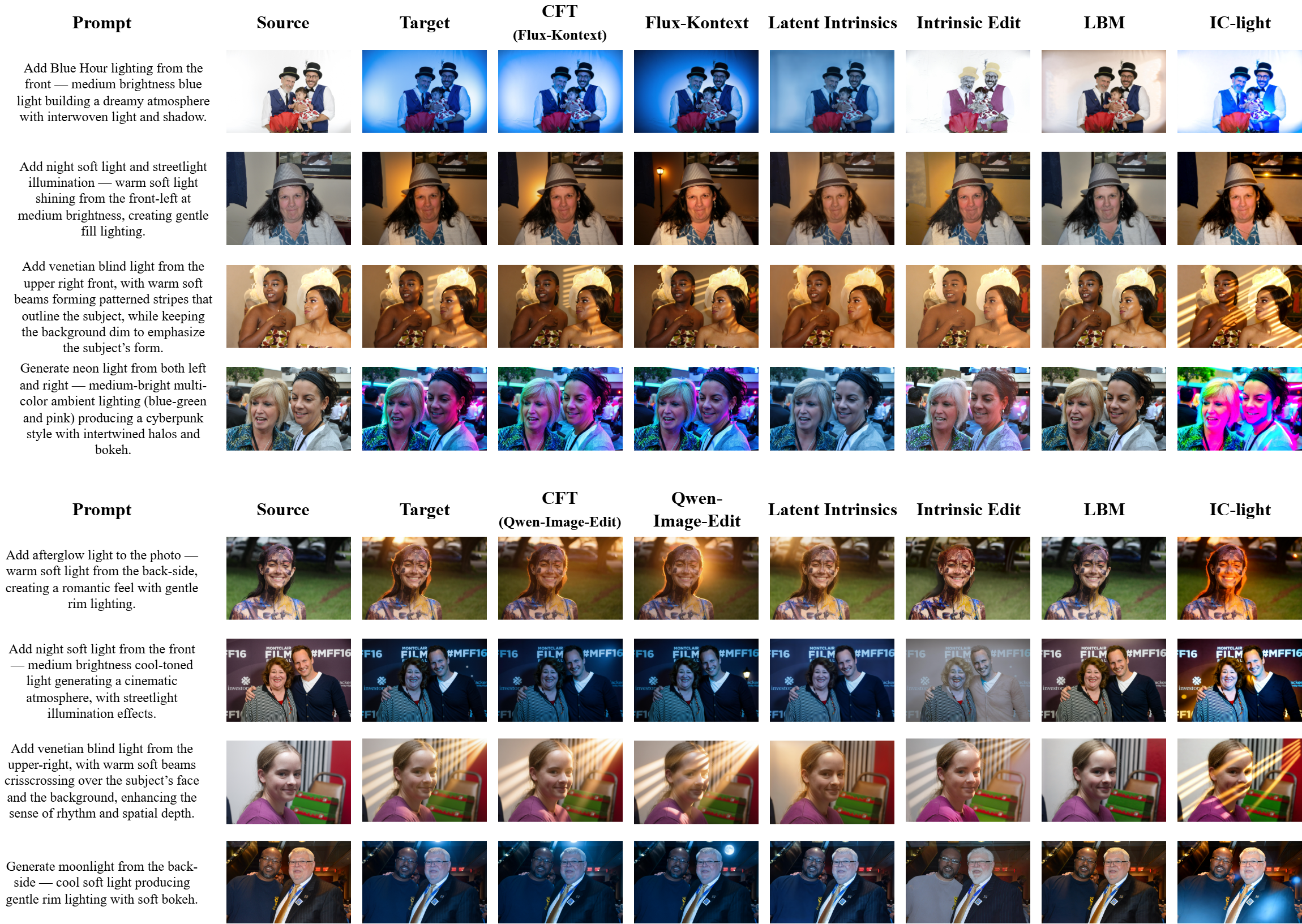}
    \caption{Qualitative comparison results.}
    \vspace{-1em}
    \label{fig:relighting_visualization}
\end{figure}


\Cref{fig:relighting_visualization} presents qualitative comparison results under multiple lighting effects. 
These examples involve varying degrees of spatially structured illumination, color shifts, and multi-source interactions, providing a comprehensive evaluation of relighting capability.
Compared with the respective base models (Flux-Kontext and Qwen-Image-Edit), CFT produces more consistent and physically coherent lighting transformations. 
For other representative baseline methods, IC-light can follow prompt instructions, but may introduce color artifacts and limited physical coherence.
Intrinsic Edit and Latent Intrinsic sometimes fail to preserve content fidelity or follow the editing descriptions, likely due to the inaccurate intrinsic decomposition and editing. 
LBM, while capable of preserving the original content of the image, often produces a rather subtle relighting effect, especially for those images with a relatively complex structure.
Overall, the qualitative results demonstrate that the proposed consistent feature transport mechanism improves illumination realism, spatial coherence, and non-lighting content preservation across diverse and challenging relighting scenarios. 
The improvements are particularly pronounced in complex lighting settings.

\begin{table}[t]
\caption{Transfer learning results.}
\label{tab:transfer-learning}
\centering
\begin{tabular}{cccccc}
\hline
Methods                & SSIM $\uparrow$           & PSNR $\uparrow$            & LPIPS $\downarrow$          & FID  $\downarrow$      & IE $\downarrow$                \\ \hline
Flux-Kontext (original) & 0.5431          & 13.5796          & 0.2331          & 50.7431          & 29.3547          \\
Flux-Kontext (w.o. CFT)           & 0.6924          & 14.4246          & 0.2031          & 44.2741          & 28.9650          \\ \hline
Flux-Kontext (w. CFT)      & 0.7325 & 15.0590 & 0.1917 & 43.5097 & 27.3482 \\ \hline
\end{tabular}
\end{table}

\subsection{Further Analysis}

To further demonstrate the effectiveness of our method, we conduct a transfer learning experiment on the real-world Multi-Illumination Dataset~\cite{murmann2019dataset}. 
Specifically, the Flux-Kontext model trained on our constructed dataset with CFT is directly evaluated on the test set of the Multi-Illumination Dataset without any additional fine-tuning. 
For comparison, we also evaluate two baselines under the same setting: (1) a Flux-Kontext model trained on our constructed dataset without CFT, and (2) the original pretrained Flux-Kontext model.
The results are shown in \Cref{tab:transfer-learning}. 
Our constructed dataset consistently improves model performance on previously unseen real-world data, indicating that the constructed dataset contains meaningful lighting features that generalize well to real-world data.
Furthermore, applying CFT leads to additional performance gains, suggesting that CFT enables the model to better capture lighting-related features and transfer the learned lighting knowledge across datasets.

\begin{table}[t]
\caption{User study results.}
\label{tab:user_study}
\centering
\begin{tabular}{ccccc}
\hline
Method                     & IQ $\uparrow$    & CC  $\uparrow$   & PP  $\uparrow$   & PA  $\uparrow$   \\ \hline
IC-light                   & 2.0636 & 2.2227 & 2.0136 & 2.0091 \\
LBM                        & 1.9864 & 2.5455 & 2.3864 & 2.2682 \\
IntrinsicEdit              & 2.1864 & 2.2136 & 2.2727 & 2.2045 \\
Latent Intrinsics         & 2.5091 & 2.9227 & 2.8227 & 2.7045 \\
Qwen-Image-Edit (w.o. CFT) & 3.5818 & 3.8091 & 3.6455 & 3.4727 \\
Flux-Kontext (w.o. CFT)    & 3.5950 & 3.7025 & 3.6529 & 3.5620 \\ \hline
Qwen-Image-Edit (w. CFT)   & 4.0091 & 4.1515 & 4.1616 & 4.0101 \\
Flux-Kontext (w. CFT)      & 4.1414 & 4.3030 & 4.2828 & 4.1919 \\ \hline
\end{tabular}
\end{table}

Relighting is highly relevant to human perception.
Therefore, we conduct a user study to evaluate the generated images from a subjective perspective.
Specifically, we randomly sample 40 image pairs from the test set and invite 30 volunteers to evaluate the results produced by different methods.
Participants are asked to rate each image on a scale from 1 to 5 according to four criteria: Illumination Quality (IQ), Content Consistency (CC), Physical Plausibility (PP), and Image Aesthetics (IA). 
Table~\ref{tab:user_study} shows the results. 
Our method achieves the highest scores across all evaluation dimensions on both Flux-Kontext and Qwen-Image-Edit backbones, further validating the effectiveness of CFT from a human perceptual perspective.

\begin{table}[t]
\caption{Ablation study results.}
\label{tab:ablation-study}
\centering
\begin{tabular}{ccccc}
\hline
Methods             & SSIM $\uparrow$    & PSNR $\uparrow$        & LPIPS $\downarrow$          & FID $\downarrow$              \\ \hline
Loss1               & 0.9153          & 22.9192          & 0.0985          & 20.3817          \\
Loss1+Loss2         & 0.9141          & 22.8299          & 0.1018          & 21.6054          \\
Loss1+Loss3         & 0.9161          & 23.3778          & 0.0972          & 19.3191          \\
Original paris as gt for Loss3 & 0.9141 & 22.8031 & 0.1003 & 20.2961 \\
Random paris as gt for Loss3 & 0.9135          & 22.7958          & 0.1035          & 19.5538          \\ \hline
CFT                 & 0.9202 & 23.5105 & 0.0946 & 18.7338 \\ \hline
\end{tabular}
\end{table}

\begin{figure}[t]
    \centering
   \begin{subfigure}{0.49\textwidth} 
    \includegraphics[width=0.9\linewidth]{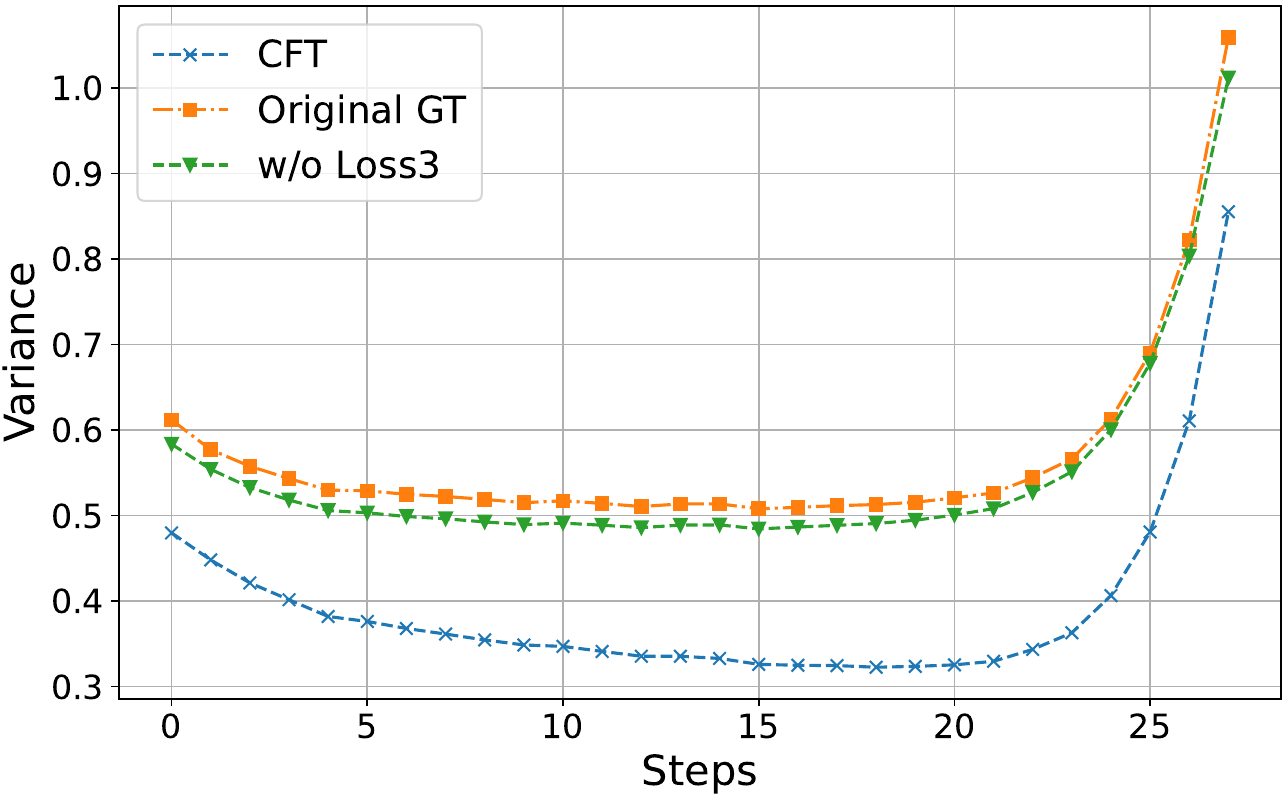}
    \caption{trainset}
    \label{fig:variance_train}
  \end{subfigure}
  \hfill
  \begin{subfigure}{0.49\textwidth} 
     \includegraphics[width=0.9\linewidth]{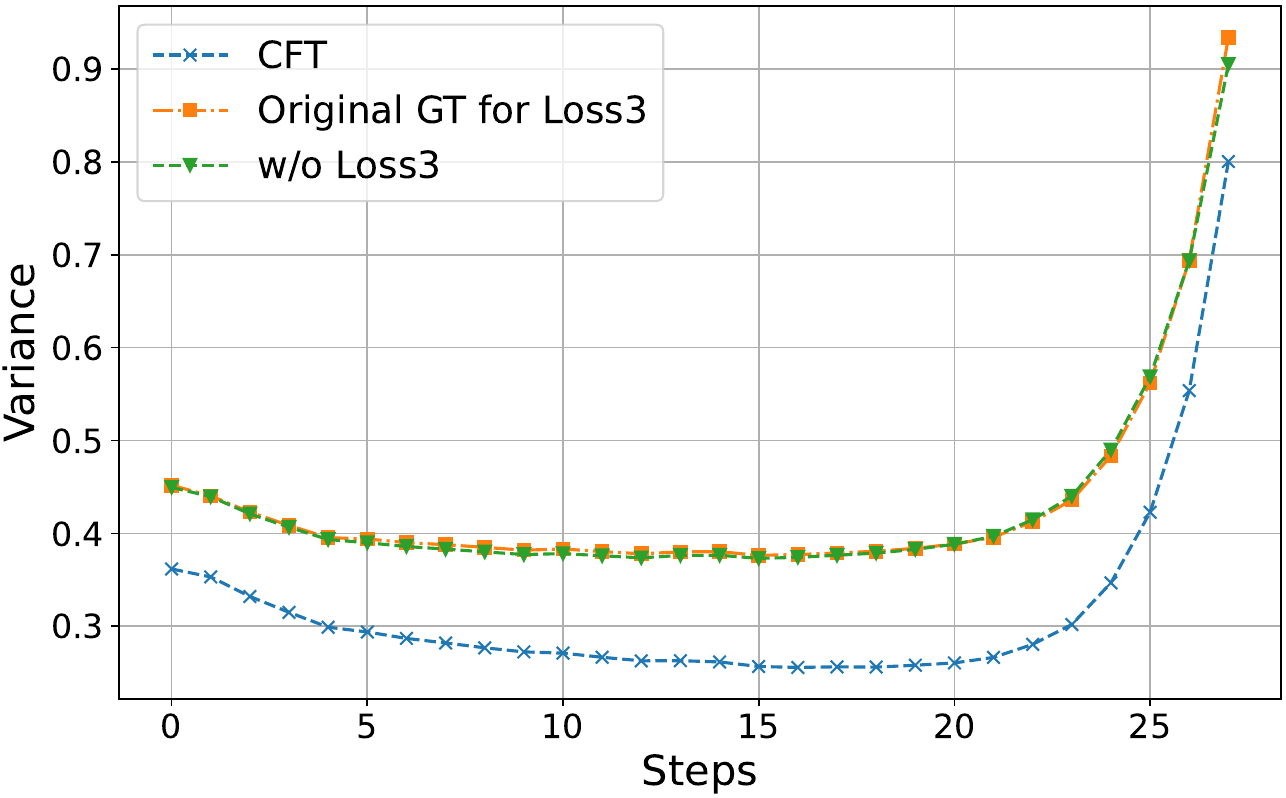}
    \caption{testset}
    \label{fig:variance_test}
  \end{subfigure}
  \caption{Variance of the predicted velocity.}
  \vspace{-1em}
  \label{fig:variance}
\end{figure}

We conduct an ablation study to verify the necessity of each component in the CFT framework.
\Cref{tab:ablation-study} presents the ablation results.
The baseline using only $\mathcal{L}_1$ corresponds to standard rectified flow training. 
Only adding $\mathcal{L}_2$, the noise-to-source reconstruction, does not improve performance and slightly degrades across all metrics, indicating that enforcing source reconstruction alone is insufficient to enhance relighting quality.
In contrast, incorporating $\mathcal{L}_3$ together with $\mathcal{L}_1$ leads to consistent improvements across all metrics, demonstrating that modeling the direct source-to-target transport is beneficial for capturing illumination-consistent features.
However, compared with the full CFT formulation, the objects without $\mathcal{L}_2$ still show inferior performance, highlighting the necessity of introducing $\mathcal{L}_2$ on top of $\mathcal{L}_1+\mathcal{L}_3$. 
By encouraging more accurate reconstruction of the source image, $\mathcal{L}_2$ provides additional structural preservation that helps the model better learn illumination feature transport between the source and target domains.

To further validate the effectiveness of the feature transport between the source and target image distribution, we conduct additional ablation experiments focusing on the supervision of $\mathcal{L}_3$. 
In our formulation, the supervision for $\mathcal{L}_3$ is constructed from another image pair $(x_{\mathrm{src}}', x_{\mathrm{tgt}}')$ sharing the same illumination transformation. 
In this experiment, we replace the ground truth of $\mathcal{L}_3$ with (1) the original source-target image pair $(x_{\mathrm{src}}, x_{\mathrm{tgt}})$ and (2) randomly sampled image pairs $(x_{\mathrm{src}}^{rand}, x_{\mathrm{tgt}}^{rand})$. 
\Cref{tab:ablation-study} shows the results.
We observe that both alternative ground truths lead to inferior performance, and even underperform the baseline trained with $\mathcal{L}_1$ alone. 
Using the original source–target image pair $(x_{\mathrm{src}}, x_{\mathrm{tgt}})$ only encourages the model to learn instance-level transformations, which is less effective for learning meaningful feature transformations.
Randomly sampled pairs $(x_{\mathrm{src}}^{rand}, x_{\mathrm{tgt}}^{rand})$ fail to provide meaningful cues for the relighting transformation.
In contrast, CFT leverages image pairs sharing the same illumination transformation as supervision, which significantly improves all evaluation metrics.
This result validates the effectiveness of modeling illumination feature transport across instances.

\begin{figure}[t]
    \centering
   \begin{subfigure}{0.24\textwidth} 
    \includegraphics[width=\linewidth]{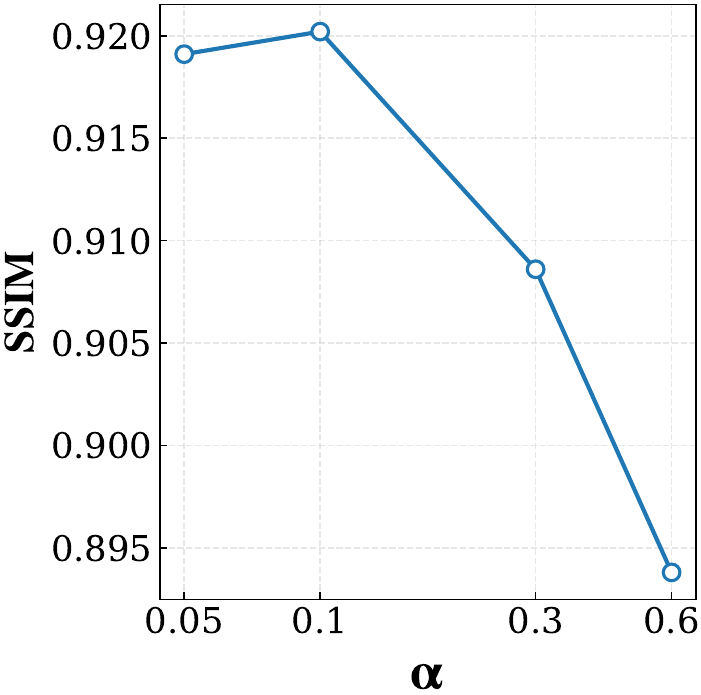}
    \caption{SSIM}
    \label{fig:fig_ssim}
  \end{subfigure}
  \hfill
  \begin{subfigure}{0.24\textwidth} 
     \includegraphics[width=\linewidth]{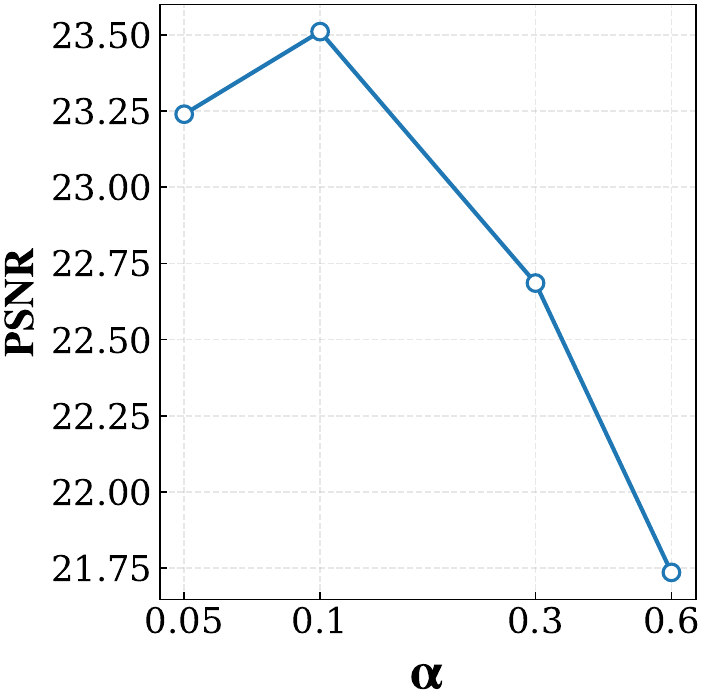}
    \caption{PSNR}
    \label{fig:fig_psnr}
  \end{subfigure}
  \hfill
  \begin{subfigure}{0.24\textwidth} 
     \includegraphics[width=\linewidth]{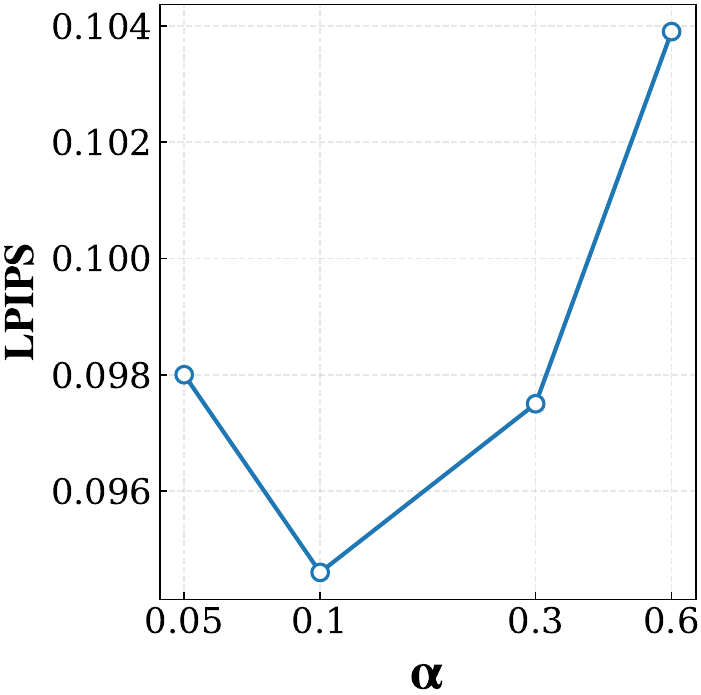}
    \caption{LPIPS}
    \label{fig:fig_lpips}
  \end{subfigure}
  \hfill
  \begin{subfigure}{0.24\textwidth} 
     \includegraphics[width=\linewidth]{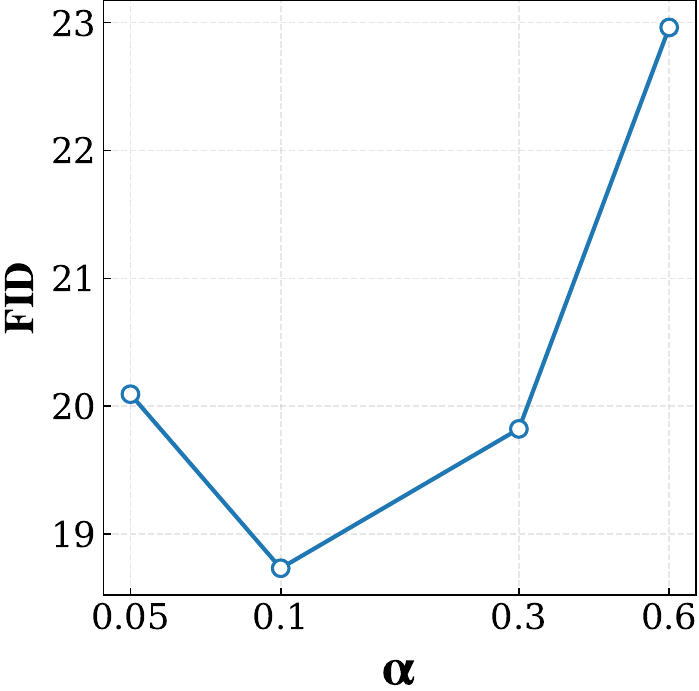}
    \caption{FID}
    \label{fig:fig_fid}
  \end{subfigure}
  \caption{The impact of $\alpha$.}
  \vspace{-1em}
  \label{fig:hyperparameter_analysis}
\end{figure}

To analyze the transport consistency, we compute the variance of the predicted source-to-target velocity $v_t^{\mathrm{direct}}$ at each step $t$ on both the training and testing sets. 
We compare three settings: (1) Only using $\mathcal{L}_1+\mathcal{L}_2$ without $\mathcal{L}_3$, (2) replacing the ground truth of $\mathcal{L}_3$ with the original image pair $(x_{\mathrm{src}}, x_{\mathrm{tgt}})$, and (3) the full CFT formulation. 
As shown in \Cref{fig:variance}, the full CFT method consistently exhibits the lowest variance. 
It means that CFT learns a more coherent and consistent feature transport field along the source-to-target transformation.


We further conduct a hyperparameter study on the weighting factor $\alpha$ that balances the feature transport loss $\mathcal{L}_3$ and the noise-to-image generation objectives $\mathcal{L}_1+\mathcal{L}_2$. 
Specifically, we evaluate $\alpha \in \{0.05, 0.1, 0.3, 0.6\}$, and the results are shown in \Cref{fig:hyperparameter_analysis}.
From the results, we observe a trade-off between the noise-to-image generation and the illumination-consistent feature transport. 
When $\alpha$ is too small (\eg, $0.05$), the influence of the transport supervision becomes limited, and the model mainly behaves like standard conditional generation, leading to weaker illumination-consistent transformations. 
In contrast, when $\alpha$ becomes too large (\eg, $0.3$ or $0.6$), the feature transport starts to dominate the optimization process and interferes with the original generative dynamics, which degrades both reconstruction fidelity and perceptual quality.
The best performance is achieved when $\alpha = 0.1$, which consistently yields the highest SSIM and PSNR as well as the lowest LPIPS and FID. 
This result indicates that a moderate transport weight provides an effective balance, allowing the model to learn illumination-consistent feature transport while preserving the stability of the underlying diffusion generation process.

\begin{table}[t]
\caption{Results for the Style Transfer task.}
\label{tab:style-transfer}
\centering
\begin{tabular}{ccccc}
\hline
Models               & SSIM $\uparrow$    & PSNR $\uparrow$        & LPIPS $\downarrow$          & FID $\downarrow$              \\ \hline
QmniStyle (w.o. CFT)           & 0.5222          & 13.804           & 0.4125          & 82.2332          \\
Qwen-Image-Edit (w.o. CFT)     & 0.4422          & 12.9693          & 0.3697          & 78.7016 \\ \hline
OmniStyle (w. CFT)       & 0.5275 & 14.1224 & 0.3945          & 85.6465          \\
Qwen-Image-Edit (w. CFT) & 0.4648          & 13.3707          & 0.3562 & 81.5195          \\ \hline
\end{tabular}
\end{table}

\subsection{Generalizing to the Style Transfer Task}
While our method is designed for image relighting, the proposed consistent feature transport formulation is not limited to relighting editing and can be applied to other image editing tasks. 
To evaluate its generality, we conduct additional experiments on image style transfer. 
We construct a style transfer dataset based on OmniStyle-150k~\cite{wang2025omnistyle} by randomly selecting 100 style categories. 
The resulting dataset contains 10{,}327 training samples and 500 testing samples. 
Each sample is organized as a triplet $(x_{\mathrm{src}}, x_{\mathrm{style}}, x_{\mathrm{tgt}})$, where $x_{\mathrm{src}}$ is the content source image, $x_{\mathrm{style}}$ is the reference style image, and $x_{\mathrm{tgt}}$ is the corresponding stylized target image.
For evaluation, we adopt OmniStyle and Qwen-Image-Edit-2509 as base models. 
We apply the proposed CFT method to both backbones under the same training settings as in the relighting task, except that the batch size is set to 2.


\Cref{tab:style-transfer} reports the results for the style transfer task. 
Applying CFT consistently improves performance across different base models.
Specifically, CFT achieves higher PSNR and SSIM and lower LPIPS on both OmniStyle and Qwen-Image-Edit, indicating that explicitly modeling feature transport helps the model better preserve content structure and improve perceptual fidelity while applying style-related transformations.
However, the FID scores of the CFT results are slightly worse than those of the original baselines.
This suggests a trade-off between transport consistency and stylistic diversity. 
Since style transfer is inherently a one-to-many generation task, where multiple stylizations may be equally plausible for the same source image, the consistent transport supervision in CFT encourages more deterministic feature transport and stronger structural alignment between source and target pairs. 
While this improves reconstruction fidelity and perceptual consistency, it may slightly reduce the global distribution diversity. 
Overall, the above results validate that the idea of consistent feature transport can be extended to other image editing tasks, such as style transfer.

\begin{figure}[t]
    \centering
    \includegraphics[width=\linewidth]{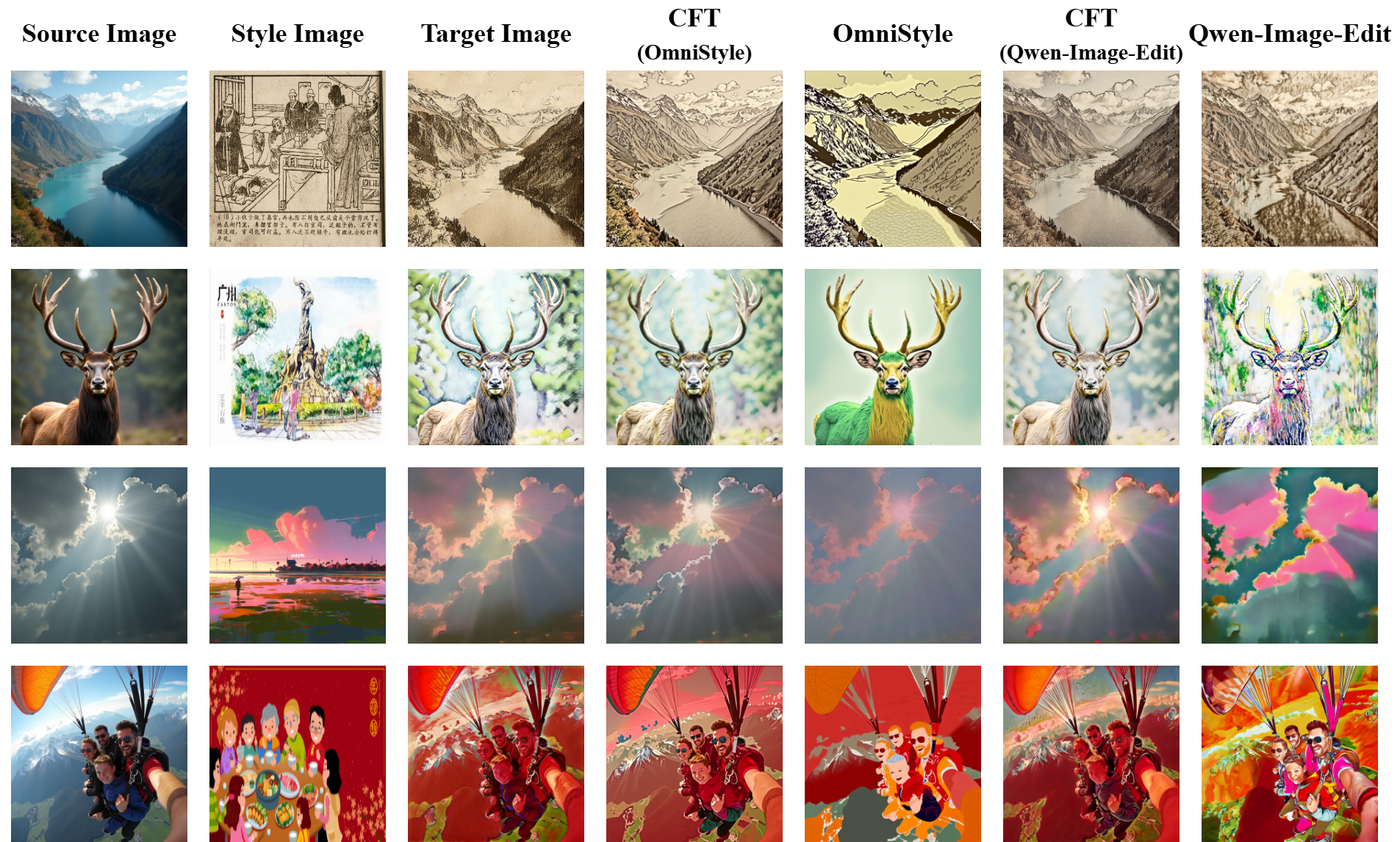}
    \caption{Qualitative comparison results for the style transfer task.}
    \vspace{-1em}
    \label{fig:style_transfer_visualzation}
\end{figure}

We present a qualitative comparison for the style transfer task in \Cref{fig:style_transfer_visualzation}.
As shown in the figure, the original base models often produce style transfer results that are either overly stylized or insufficiently stylized. 
By employing the proposed consistent feature transport (CFT), the generated images exhibit a more balanced style transformation. 
Our method better preserves the structural content of the source image while transferring the key stylistic patterns from the reference style image, resulting in visually coherent and stylistically faithful outputs across different scenes and styles.
\section{Conclusion}

In this work, we propose a Consistent Feature Transport (CFT) method for image relighting. 
By jointly modeling noise-to-image generation and illumination-consistent source-to-target transport, CFT enables the model to focus on illumination feature transformation while preserving non-lighting content. 
In addition, we construct a portrait relighting dataset with complex illumination effects, addressing the scarcity of suitable data resources. 
Extensive experiments demonstrate that the CFT achieves consistent improvements over existing relighting approaches in both quantitative and qualitative evaluations. 
Furthermore, experiments on style transfer show that the proposed feature transport mechanism generalizes beyond relighting and can be applied to broader image editing tasks.
In the future, we plan to apply CFT to additional editing scenarios, exploring its effectiveness for modeling feature transport across different editing tasks. 

\section*{Acknowledgements}
This work was supported by the General Program of Beijing Natural Science Foundation under Grant 4262030.

%
%
\bibliographystyle{splncs04}
\bibliography{main}

@inproceedings{kim2024switchlight,
  title={Switchlight: Co-design of physics-driven architecture and pre-training framework for human portrait relighting},
  author={Kim, Hoon and Jang, Minje and Yoon, Wonjun and Lee, Jisoo and Na, Donghyun and Woo, Sanghyun},
  booktitle={Proceedings of the IEEE/CVF Conference on Computer Vision and Pattern Recognition},
  pages={25096--25106},
  year={2024}
}

@inproceedings{zhang2025scaling,
  title={Scaling in-the-wild training for diffusion-based illumination harmonization and editing by imposing consistent light transport},
  author={Zhang, Lvmin and Rao, Anyi and Agrawala, Maneesh},
  booktitle={The Thirteenth International Conference on Learning Representations},
  year={2025}
}

@inproceedings{magar2025lightlab,
  title={Lightlab: Controlling light sources in images with diffusion models},
  author={Magar, Nadav and Hertz, Amir and Tabellion, Eric and Pritch, Yael and Rav-Acha, Alex and Shamir, Ariel and Hoshen, Yedid},
  booktitle={Proceedings of the Special Interest Group on Computer Graphics and Interactive Techniques Conference Conference Papers},
  pages={1--11},
  year={2025}
}

@inproceedings{kocsis2024lightit,
  title={Lightit: Illumination modeling and control for diffusion models},
  author={Kocsis, Peter and Philip, Julien and Sunkavalli, Kalyan and Nie{\ss}ner, Matthias and Hold-Geoffroy, Yannick},
  booktitle={Proceedings of the IEEE/CVF Conference on Computer Vision and Pattern Recognition},
  pages={9359--9369},
  year={2024}
}

@article{zhang2024latent,
  title={Latent intrinsics emerge from training to relight},
  author={Zhang, Xiao and Gao, William and Jain, Seemandhar and Maire, Michael and Forsyth, David and Bhattad, Anand},
  journal={Advances in Neural Information Processing Systems},
  volume={37},
  pages={96775--96796},
  year={2024}
}

@inproceedings{chadebec2025lbm,
  title={Lbm: Latent bridge matching for fast image-to-image translation},
  author={Chadebec, Cl{\'e}ment and Tasar, Onur and Sreetharan, Sanjeev and Aubin, Benjamin},
  booktitle={Proceedings of the IEEE/CVF International Conference on Computer Vision},
  pages={29086--29098},
  year={2025}
}

@inproceedings{murmann2019dataset,
  title={A dataset of multi-illumination images in the wild},
  author={Murmann, Lukas and Gharbi, Michael and Aittala, Miika and Durand, Fredo},
  booktitle={Proceedings of the IEEE/CVF International Conference on Computer Vision},
  pages={4080--4089},
  year={2019}
}

@inproceedings{hold2019deep,
  title={Deep sky modeling for single image outdoor lighting estimation},
  author={Hold-Geoffroy, Yannick and Athawale, Akshaya and Lalonde, Jean-Fran{\c{c}}ois},
  booktitle={Proceedings of the IEEE/CVF conference on computer vision and pattern recognition},
  pages={6927--6935},
  year={2019}
}

@inproceedings{brooks2023instructpix2pix,
  title={Instructpix2pix: Learning to follow image editing instructions},
  author={Brooks, Tim and Holynski, Aleksander and Efros, Alexei A},
  booktitle={Proceedings of the IEEE/CVF conference on computer vision and pattern recognition},
  pages={18392--18402},
  year={2023}
}

@inproceedings{wang2025omnistyle,
  title={Omnistyle: Filtering high quality style transfer data at scale},
  author={Wang, Ye and Liu, Ruiqi and Lin, Jiang and Liu, Fei and Yi, Zili and Wang, Yilin and Ma, Rui},
  booktitle={Proceedings of the Computer Vision and Pattern Recognition Conference},
  pages={7847--7856},
  year={2025}
}

@inproceedings{kulikov2025flowedit,
  title={Flowedit: Inversion-free text-based editing using pre-trained flow models},
  author={Kulikov, Vladimir and Kleiner, Matan and Huberman-Spiegelglas, Inbar and Michaeli, Tomer},
  booktitle={Proceedings of the IEEE/CVF International Conference on Computer Vision},
  pages={19721--19730},
  year={2025}
}

@article{pandey2021total,
  title={Total relighting: learning to relight portraits for background replacement.},
  author={Pandey, Rohit and Orts-Escolano, Sergio and Legendre, Chloe and Haene, Christian and Bouaziz, Sofien and Rhemann, Christoph and Debevec, Paul E and Fanello, Sean Ryan},
  journal={ACM Trans. Graph.},
  volume={40},
  number={4},
  pages={43--1},
  year={2021}
}

@inproceedings{
liu2023flow,
title={Flow Straight and Fast: Learning to Generate and Transfer Data with Rectified Flow},
author={Xingchao Liu and Chengyue Gong and qiang liu},
booktitle={The Eleventh International Conference on Learning Representations },
year={2023},
}

@article{wu2025qwen,
  title={Qwen-image technical report},
  author={Wu, Chenfei and Li, Jiahao and Zhou, Jingren and Lin, Junyang and Gao, Kaiyuan and Yan, Kun and Yin, Sheng-ming and Bai, Shuai and Xu, Xiao and Chen, Yilei and others},
  journal={arXiv preprint arXiv:2508.02324},
  year={2025}
}

@article{labs2025flux,
  title={FLUX. 1 Kontext: Flow Matching for In-Context Image Generation and Editing in Latent Space},
  author={Labs, Black Forest and Batifol, Stephen and Blattmann, Andreas and Boesel, Frederic and Consul, Saksham and Diagne, Cyril and Dockhorn, Tim and English, Jack and English, Zion and Esser, Patrick and others},
  journal={arXiv preprint arXiv:2506.15742},
  year={2025}
}

@article{lyu2025intrinsicedit,
  title={IntrinsicEdit: Precise generative image manipulation in intrinsic space},
  author={Lyu, Linjie and Deschaintre, Valentin and Hold-Geoffroy, Yannick and Ha{\v{s}}an, Milo{\v{s}} and Yoon, Jae Shin and Leimk{\"u}hler, Thomas and Theobalt, Christian and Georgiev, Iliyan},
  journal={ACM Transactions on Graphics (TOG)},
  volume={44},
  number={4},
  pages={1--13},
  year={2025},
  publisher={ACM New York, NY, USA}
}

@inproceedings{chaturvedi2025synthlight,
  title={Synthlight: Portrait relighting with diffusion model by learning to re-render synthetic faces},
  author={Chaturvedi, Sumit and Ren, Mengwei and Hold-Geoffroy, Yannick and Liu, Jingyuan and Dorsey, Julie and Shu, Zhixin},
  booktitle={Proceedings of the IEEE/CVF Conference on Computer Vision and Pattern Recognition},
  pages={369--379},
  year={2025}
}

@inproceedings{ren2024relightful,
  title={Relightful harmonization: Lighting-aware portrait background replacement},
  author={Ren, Mengwei and Xiong, Wei and Yoon, Jae Shin and Shu, Zhixin and Zhang, Jianming and Jung, HyunJoon and Gerig, Guido and Zhang, He},
  booktitle={Proceedings of the IEEE/CVF Conference on Computer Vision and Pattern Recognition},
  pages={6452--6462},
  year={2024}
}

@article{li2025translight,
  title={TransLight: Image-Guided Customized Lighting Control with Generative Decoupling},
  author={Li, Zongming and Zhu, Lianghui and Shen, Haocheng and Ran, Longjin and Liu, Wenyu and Wang, Xinggang},
  journal={arXiv preprint arXiv:2508.14814},
  year={2025}
}

@article{liu2025dreamlight,
  title={DreamLight: Towards Harmonious and Consistent Image Relighting},
  author={Liu, Yong and Xiao, Wenpeng and Wang, Qianqian and Chen, Junlin and Wang, Shiyin and Wang, Yitong and Wu, Xinglong and Tang, Yansong},
  journal={arXiv preprint arXiv:2506.14549},
  year={2025}
}

@article{liu2023openillumination,
  title={Openillumination: A multi-illumination dataset for inverse rendering evaluation on real objects},
  author={Liu, Isabella and Chen, Linghao and Fu, Ziyang and Wu, Liwen and Jin, Haian and Li, Zhong and Wong, Chin Ming Ryan and Xu, Yi and Ramamoorthi, Ravi and Xu, Zexiang and others},
  journal={Advances in Neural Information Processing Systems},
  volume={36},
  pages={36951--36962},
  year={2023}
}

@article{helou2020vidit,
  title={VIDIT: Virtual image dataset for illumination transfer},
  author={Helou, Majed El and Zhou, Ruofan and Barthas, Johan and S{\"u}sstrunk, Sabine},
  journal={arXiv preprint arXiv:2005.05460},
  year={2020}
}

@inproceedings{xing2025luminet,
  title={Luminet: Latent intrinsics meets diffusion models for indoor scene relighting},
  author={Xing, Xiaoyan and Groh, Konrad and Karaoglu, Sezer and Gevers, Theo and Bhattad, Anand},
  booktitle={Proceedings of the Computer Vision and Pattern Recognition Conference},
  pages={442--452},
  year={2025}
}

@inproceedings{choi2025scribblelight,
  title={Scribblelight: Single image indoor relighting with scribbles},
  author={Choi, Jun Myeong and Wang, Annie and Peers, Pieter and Bhattad, Anand and Sengupta, Roni},
  booktitle={Proceedings of the Computer Vision and Pattern Recognition Conference},
  pages={5720--5731},
  year={2025}
}

@article{zhang2023magicbrush,
  title={Magicbrush: A manually annotated dataset for instruction-guided image editing},
  author={Zhang, Kai and Mo, Lingbo and Chen, Wenhu and Sun, Huan and Su, Yu},
  journal={Advances in Neural Information Processing Systems},
  volume={36},
  pages={31428--31449},
  year={2023}
}

@article{yang2023imagebrush,
  title={Imagebrush: Learning visual in-context instructions for exemplar-based image manipulation},
  author={Yang, Yifan and Peng, Houwen and Shen, Yifei and Yang, Yuqing and Hu, Han and Qiu, Lili and Koike, Hideki and others},
  journal={Advances in Neural Information Processing Systems},
  volume={36},
  pages={48723--48743},
  year={2023}
}

@inproceedings{dalva2025fluxspace,
  title={FluxSpace: Disentangled Semantic Editing in Rectified Flow Models},
  author={Dalva, Yusuf and Venkatesh, Kavana and Yanardag, Pinar},
  booktitle={Proceedings of the Computer Vision and Pattern Recognition Conference},
  pages={13083--13092},
  year={2025}
}

@inproceedings{pan2025counterfactual,
  title={Counterfactual image editing with disentangled causal latent space},
  author={Pan, Yushu and Bareinboim, Elias},
  booktitle={The Thirty-ninth Annual Conference on Neural Information Processing Systems},
  year={2025}
}

@inproceedings{zhang2024choose,
  title={Choose what you need: Disentangled representation learning for scene text recognition removal and editing},
  author={Zhang, Boqiang and Xie, Hongtao and Gao, Zuan and Wang, Yuxin},
  booktitle={Proceedings of the IEEE/CVF conference on computer vision and pattern recognition},
  pages={28358--28368},
  year={2024}
}

@article{tan2025vision,
  title={Vision Bridge Transformer at Scale},
  author={Tan, Zhenxiong and Wang, Zeqing and Yang, Xingyi and Liu, Songhua and Wang, Xinchao},
  journal={arXiv preprint arXiv:2511.23199},
  year={2025}
}

@inproceedings{
wang2025taming,
title={Taming Rectified Flow for Inversion and Editing},
author={Jiangshan Wang and Junfu Pu and Zhongang Qi and Jiayi Guo and Yue Ma and Nisha Huang and Yuxin Chen and Xiu Li and Ying Shan},
booktitle={Forty-second International Conference on Machine Learning},
year={2025}
}

@inproceedings{
deng2025fireflow,
title={FireFlow: Fast Inversion of Rectified Flow for Image Semantic Editing},
author={Yingying Deng and Xiangyu He and Changwang Mei and Peisong Wang and Fan Tang},
booktitle={Forty-second International Conference on Machine Learning},
year={2025}
}

@inproceedings{
kim2026flowalign,
title={FlowAlign: Trajectory-Regularized, Inversion-Free Flow-based Image Editing},
author={Jeongsol Kim and Yeobin Hong and Jonghyun Park and Jong Chul Ye},
booktitle={The Fourteenth International Conference on Learning Representations},
year={2026}
}

@inproceedings{li2022physically,
  title={Physically-based editing of indoor scene lighting from a single image},
  author={Li, Zhengqin and Shi, Jia and Bi, Sai and Zhu, Rui and Sunkavalli, Kalyan and Ha{\v{s}}an, Milo{\v{s}} and Xu, Zexiang and Ramamoorthi, Ravi and Chandraker, Manmohan},
  booktitle={European Conference on Computer Vision},
  pages={555--572},
  year={2022},
  organization={Springer}
}

@inproceedings{yu2020self,
  title={Self-supervised outdoor scene relighting},
  author={Yu, Ye and Meka, Abhimitra and Elgharib, Mohamed and Seidel, Hans-Peter and Theobalt, Christian and Smith, William AP},
  booktitle={European Conference on Computer Vision},
  pages={84--101},
  year={2020},
  organization={Springer}
}

@inproceedings{kocsis2024intrinsic,
  title={Intrinsic image diffusion for indoor single-view material estimation},
  author={Kocsis, Peter and Sitzmann, Vincent and Nie{\ss}ner, Matthias},
  booktitle={Proceedings of the IEEE/CVF Conference on Computer Vision and Pattern Recognition},
  pages={5198--5208},
  year={2024}
}

@inproceedings{choi2024improving,
  title={Improving diffusion models for authentic virtual try-on in the wild},
  author={Choi, Yisol and Kwak, Sangkyung and Lee, Kyungmin and Choi, Hyungwon and Shin, Jinwoo},
  booktitle={European Conference on Computer Vision},
  pages={206--235},
  year={2024},
  organization={Springer}
}

@inproceedings{cao2023masactrl,
  title={Masactrl: Tuning-free mutual self-attention control for consistent image synthesis and editing},
  author={Cao, Mingdeng and Wang, Xintao and Qi, Zhongang and Shan, Ying and Qie, Xiaohu and Zheng, Yinqiang},
  booktitle={Proceedings of the IEEE/CVF international conference on computer vision},
  pages={22560--22570},
  year={2023}
}

@inproceedings{karras2019style,
  title={A style-based generator architecture for generative adversarial networks},
  author={Karras, Tero and Laine, Samuli and Aila, Timo},
  booktitle={Proceedings of the IEEE/CVF conference on computer vision and pattern recognition},
  pages={4401--4410},
  year={2019}
}

@article{kalantari2017deep,
  title={Deep high dynamic range imaging of dynamic scenes.},
  author={Kalantari, Nima Khademi and Ramamoorthi, Ravi and others},
  journal={ACM Trans. Graph.},
  volume={36},
  number={4},
  pages={144--1},
  year={2017}
}

@article{hasinoff2016burst,
  title={Burst photography for high dynamic range and low-light imaging on mobile cameras},
  author={Hasinoff, Samuel W and Sharlet, Dillon and Geiss, Ryan and Adams, Andrew and Barron, Jonathan T and Kainz, Florian and Chen, Jiawen and Levoy, Marc},
  journal={ACM Transactions on Graphics (ToG)},
  volume={35},
  number={6},
  pages={1--12},
  year={2016},
  publisher={ACM New York, NY, USA}
}

@article{bai2025qwen3,
  title={Qwen3-vl technical report},
  author={Bai, Shuai and Cai, Yuxuan and Chen, Ruizhe and Chen, Keqin and Chen, Xionghui and Cheng, Zesen and Deng, Lianghao and Ding, Wei and Gao, Chang and Ge, Chunjiang and others},
  journal={arXiv preprint arXiv:2511.21631},
  year={2025}
}

@inproceedings{
luo2026editscore,
title={EditScore: Unlocking Online {RL} for Image Editing via High-Fidelity Reward Modeling},
author={Xin Luo and Jiahao Wang and Chenyuan Wu and Shitao Xiao and Xiyan Jiang and Defu Lian and Jiajun Zhang and Dong Liu and Zheng Liu},
booktitle={The Fourteenth International Conference on Learning Representations},
year={2026}
}

@article{hurst2024gpt,
  title={Gpt-4o system card},
  author={Hurst, Aaron and Lerer, Adam and Goucher, Adam P and Perelman, Adam and Ramesh, Aditya and Clark, Aidan and Ostrow, AJ and Welihinda, Akila and Hayes, Alan and Radford, Alec and others},
  journal={arXiv preprint arXiv:2410.21276},
  year={2024}
}

@inproceedings{debevec2000acquiring,
  title={Acquiring the reflectance field of a human face},
  author={Debevec, Paul and Hawkins, Tim and Tchou, Chris and Duiker, Haarm-Pieter and Sarokin, Westley and Sagar, Mark},
  booktitle={Proceedings of the 27th annual conference on Computer graphics and interactive techniques},
  pages={145--156},
  year={2000}
}

@inproceedings{wang2025comprehensive,
  title={Comprehensive relighting: Generalizable and consistent monocular human relighting and harmonization},
  author={Wang, Junying and Liu, Jingyuan and Sun, Xin and Singh, Krishna Kumar and Shu, Zhixin and Zhang, He and Yang, Jimei and Zhao, Nanxuan and Wang, Tuanfeng Y and Chen, Simon S and others},
  booktitle={Proceedings of the Computer Vision and Pattern Recognition Conference},
  pages={380--390},
  year={2025}
}

@article{liu2026unilumos,
  title={Unilumos: Fast and unified image and video relighting with physics-plausible feedback},
  author={Liu, Pengwei and Yuan, Hangjie and Dong, Bo and Xing, Jiazheng and Wang, Jinwang and Zhao, Rui and Chen, Weihua and Wang, Fan},
  journal={Advances in Neural Information Processing Systems},
  volume={38},
  pages={82052--82080},
  year={2026}
}

@inproceedings{bharadwaj2025genlit,
  title={Genlit: Reformulating single-image relighting as video generation},
  author={Bharadwaj, Shrisha and Feng, Haiwen and Becherini, Giorgio and Fernandez Abrevaya, Victoria and Black, Michael J},
  booktitle={Proceedings of the SIGGRAPH Asia 2025 Conference Papers},
  pages={1--12},
  year={2025}
}

@article{chen2026car,
  title={CAR-Flow: Condition-Aware Reparameterization Aligns Source and Target for Better Flow Matching},
  author={Chen, Chen and Guo, Pengsheng and Song, Liangchen and Lu, Jiasen and Qian, Rui and Fu, Tsu-Jui and Wang, Xinze and Liu, Wei and Yang, Yinfei and Schwing, Alex},
  journal={Advances in Neural Information Processing Systems},
  volume={38},
  pages={94919--94945},
  year={2026}
}

@inproceedings{
lipman2023flow,
title={Flow Matching for Generative Modeling},
author={Yaron Lipman and Ricky T. Q. Chen and Heli Ben-Hamu and Maximilian Nickel and Matthew Le},
booktitle={The Eleventh International Conference on Learning Representations },
year={2023}
}
\end{document}